\documentclass{article}

    \usepackage[preprint, nonatbib]{neurips_2020}
    
\usepackage[utf8]{inputenc} 
\usepackage[T1]{fontenc}    
\usepackage{hyperref}       
\usepackage{url}            
\usepackage{booktabs}       
\usepackage{amsfonts}       
\usepackage{nicefrac}       
\usepackage{microtype}      
\usepackage{biblatex}
\usepackage{graphicx}

\usepackage{floatrow}
\newfloatcommand{capbtabbox}{table}[][\FBwidth]

\title{Differentiable Histogram with Hard-Binning}

\author{%
  Ibrahim Yusuf \\
  InstaDeep\\
  Lagos, Nigeria \\
  \texttt{i.yusuf@instadeep.com} \\
   \And
   George Igwegbe \\
   InstaDeep \\
   Lagos, Nigeria \\
   \texttt{gigwegbe@gmail.com} \\
   \And
   Oluwafemi Azeez  \\
   InstaDeep \\
   Lagos, Nigeria \\
   \texttt{f.azeez@instadeep.com} \\
}

\begin{document}

\maketitle

\begin{abstract}
  The simplicity and expressiveness of a histogram render it a useful feature in different contexts including deep learning. Although the process of computing a histogram is non-differentiable, researchers have proposed differentiable approximations, which have some limitations. A differentiable histogram that directly approximates the hard-binning operation in conventional histograms is proposed. It combines the strength of existing differentiable histograms and overcomes their individual challenges. In comparison to a histogram computed using Numpy, the proposed histogram has an absolute approximation error of 0.000158.
\end{abstract}

\section{Introduction}

A histogram is a statistical feature that captures the distribution of sample values. It finds useful applications in a variety of domains including deep learning. Several recent works have incorporated histograms as a learnable feature in a deep neural network for tasks such as texture analysis, steganalysis, object detection, semantic segmentation, etc. Because a conventional histogram involves binning, which can be viewed as an indicator function whereby a sample returns one (1) if it falls within a bin and zero (0) otherwise. The works on learnable histograms have tried to approximate this hard-binning process with a soft-binning version. In the soft-binning version, a sample distributes its maximum vote of 1 across several bins, although it is expected to have the maximum vote in the bin it belongs to. However, the result of soft-binning operation falls short of what is expected from the computation of a histogram. A particular work that closely approximates a conventional histogram used Kernel Density Estimation with an appropriate kernel, but unlike others it cannot be implemented with existing CNN layers and it has a tuning parameter. In this work, a differentiable histogram that directly approximates the hard-binning used in conventional histogram computation is proposed. The proposed histogram has a simple functional form it can be implemented using CNN layers.

\section{Related Works}

Wang Z. et. al. \cite{wang2016learnable} incorporated histogram features into a deep convolutional neural network for object detection and semantic segmentation, resulting in networks namely HistNet-OD and HistNet-SS respectively. The networks add a learnable histogram layer on top of conventional CNNs and they report superior performance. The learnable histogram layer was modeled using a linear basis function. J. Peeples et al. \cite{peeples2020histogram} also proposed a learnable histogram layer for texture analysis. The learnable histogram layer is used to extract spatial distribution of feature values for characterizing and distinguishing textures. Avi-Aharon, M. et al. \cite{avi2020deephist} proposed a differentiable histogram for transferring the color of a source image to a generated image in the context of image-to-image translation with GANs. The differentiable histogram was modeled using Kernel Density Estimation (KDE) with the derivative of the logistic regression function as the kernel. The functional form of the histogram contains a parameter B of KDE which needs to be properly tuned for a good approximation. It also cannot be implemented using a single convolutional layer. In this work, we propose a differentiable histogram that approximates the hard-binning operation of a conventional histogram. It combines the strengths of existing differentiable histograms and overcomes the limitations of soft-binning and parameter tuning.






\section{Proposed Differentiable Histogram}


The proposed histogram layer overcomes the limitations of existing differentiable histograms by approximating the actual hard-binning operation found in conventional histograms using equation \ref{eq:1}.

\begin{equation}
    \Phi(1.01^{ \omega_k - |x_i - \mu_k |}, 1, 0) \approx \Big\{^{1\ if\ x_{i,j}\ \in\  (\mu_k - \omega_k, \mu_k + \omega_k) }_{0\ otherwise} \label{eq:1}
\end{equation}
\begin{equation}
    \Phi(x, 1, 0) \approx \Big\{^{x\ if\ x\ >\  1 }_{1\ otherwise} \label{eq:2}
\end{equation}

\begin{figure}[h]

\centering
    \includegraphics[
  width=15cm,
  height=10cm,
]{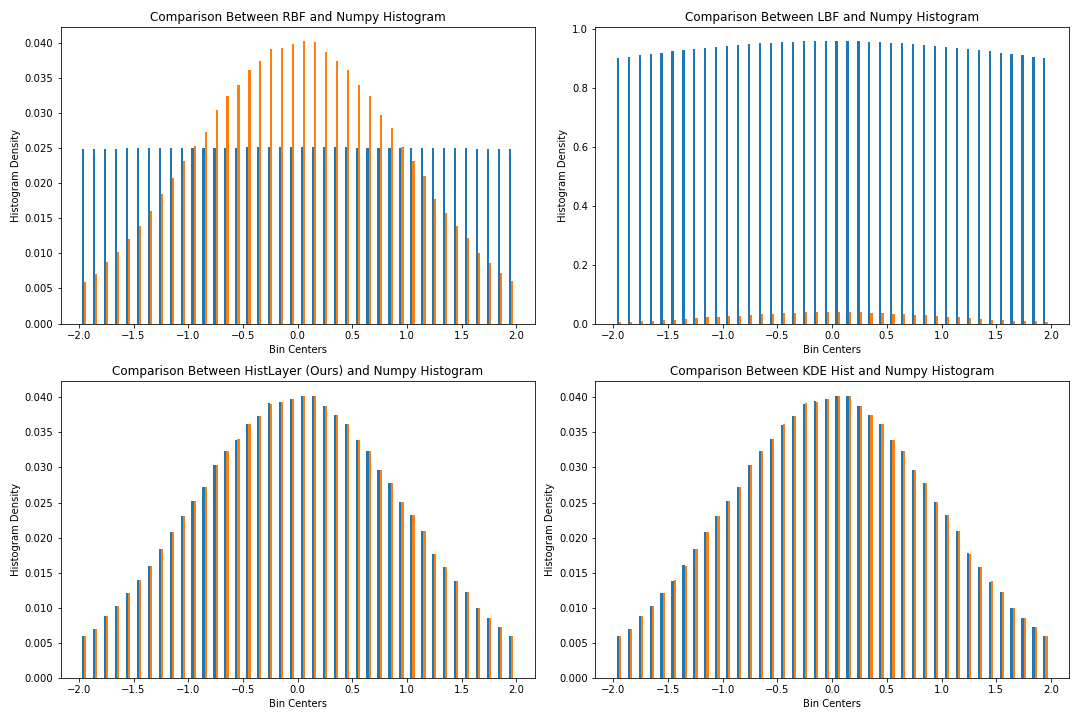}

  \caption{Histogram of a standard normal distribution with 20 equally spaced bins between -1 and 1 }%
  \label{fig:all_comp}
\end{figure}

\begin{figure}[h]
\centering
    \includegraphics[
  width=16cm,
  height=10cm,
  keepaspectratio,
]{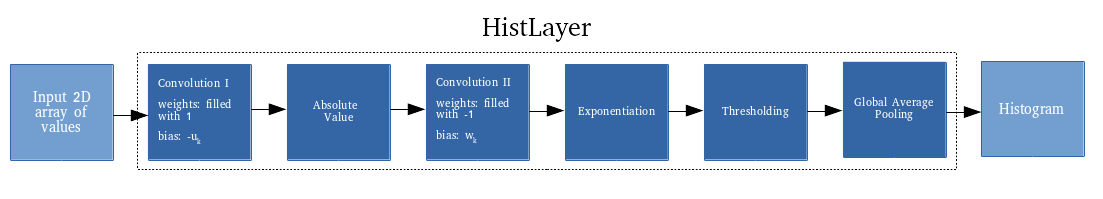}
 
  \caption{Modelling of HistLayer using convolutional layers}%
  \label{fig:hist_draw}
\end{figure}

According to equation \ref{eq:1},  $\forall x\ \in\ (\mu_k - \omega_k, \mu_k + \omega_k)$ the value of $\omega_k -|x -\mu_k| \in (0, \omega_k)$, otherwise it is  $ < 0$. Consequently, the value of $1.01^{\omega_k -|x -\mu_k|}\ \in\ (1, 1.01^{\omega_k})\ \forall x\ \in\ ((\mu_k - \omega_k, \mu_k + \omega_k)$ and  $ < 1$ otherwise. At this point, any sample outside the bin interval votes with a value strictly less than 1. To discard these unwanted votes, we simply apply a “relu at 1” using the threshold function of equation \ref{eq:2}. Fig. \ref{fig:all_comp} shows a comparison of the histogram computed using differentiable histograms and Numpy histogram function \cite{harris2020array}. Similar to \cite{peeples2020histogram} and \cite{wang2016learnable}, equation \ref{eq:1} can be easily implemented using existing CNN layers as illustrated in Fig \ref{fig:hist_draw}.

\section{Experiment}
To quantify the approximation error of equation \ref{eq:1}, we compute the absolute error between histogram extracted using equation \ref{eq:1} and that extracted in a conventional manner, using a function in Numpy \cite{harris2020array}. Table \ref{tab:tab1} shows the error for different differentiable histograms in Fig. \ref{fig:all_comp}.

\begin{figure}[h]

\centering
    \includegraphics[
  width=16cm,
  height=10cm,]{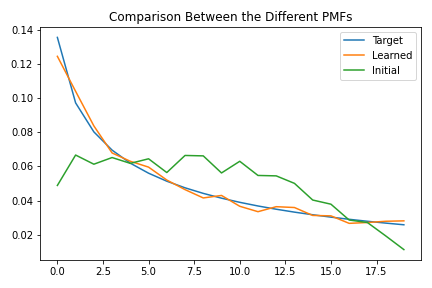}
  \caption{Histogram of DC-GAN generator output before and after learning.}%
  \label{fig:exp}
\end{figure}

\begin{figure}[H]

  \begin{center}
\begin{tabular}{ |c|c| } 
 \hline
 Differentiable Histogram & Absolute Error \\
 \hline
LBF & 36.55150 \\ 
RBF & 0.407948 \\
KDE & 0.001676\\
HistLayer (Ours) & 0.000158\\
 \hline
\end{tabular}
\end{center}
  \caption{A table of Absolute Errors of various differentiable histogram algorithms}%
  \label{tab:tab1}
\end{figure}

\section{Conclusion}
We propose a differentiable histogram layer that directly approximates the hard-binning operation in conventional histogram and show that it can be used to learn a generator that maps an input noise vector to any target distribution expressible by a histogram.

\printbibliography
\end{document}